\documentclass[12pt,a4paper]{cibb}

\usepackage{graphicx, subcaption}
\usepackage{amsmath,amsfonts,latexsym,amssymb,euscript,xr}

\usepackage{blindtext}
\usepackage{listings}
\usepackage[bookmarks=false]{hyperref}
\usepackage[utf8]{inputenc}
\usepackage{float}
\usepackage{stfloats}
\usepackage{booktabs}
\usepackage{multirow}
\usepackage[shortlabels]{enumitem}
\usepackage{hyperref}
\usepackage{flushend}
\usepackage[numbers]{natbib}

\hypersetup{
    colorlinks,
    citecolor=black,
    filecolor=black,
    linkcolor=black,
    urlcolor=black
}

\newcommand\fscore{F\textsubscript{1} score}

\title{\large $\ $\\ \bf Autoencoders as Weight Initialization of Deep Classification Networks for Cancer \emph{versus} Cancer Studies}

\author{ Mafalda Falcão Ferreira$^{*, 1, 2}$, Rui Camacho$^{1,2}$, Luís F. Teixeira$^{1,2}$}

\address{$\ $\\$^1$ Doctoral Program in Informatics Engineering of the Faculty of Engineering, University of Porto
\\ 
up201204016@fe.up.pt
\\
\bigskip
$^2$ INESC TEC - Institute for Systems and Computer Engineering, Technology and Science\\
Porto - Portugal
\\
\bigskip
$^3$ Department of Informatics Engineering - Faculty of Engineering, University of Porto\\
Porto - Portugal, \{rcamacho, luisft\}@fe.up.pt
\\
\bigskip
$^*$ Corresponding Author
}

\abstract{Cancer, Classification, Deep Learning, Autoencoders, Gene Expression Ana-lysis.
\\[17pt]
{\bf Abstract.} Cancer is still one of the most devastating diseases of our time. One way of automatically classifying tumor samples is by analyzing its derived molecular information (\emph{i.e.}, its genes expression signatures). In this work, we aim to distinguish three different types of cancer: thyroid, skin, and stomach. For that, we compare the performance of a Denoising Autoencoder (DAE) used as weight initialization of a deep neural network. Although we address a different domain problem in this work, we have adopted the same methodology of Ferreira \emph{et al.}. In our experiments, we assess two different approaches when training the classification model: (a) fixing the weights, after pre-training the DAE, and (b) allowing fine-tuning of the entire classification network. Additionally, we apply two different strategies for embedding the DAE into the classification network: (1) by only importing the encoding layers, and (2) by inserting the complete autoencoder. Our best result was the combination of unsupervised feature learning through a DAE, followed by its full import into the classification network, and subsequent fine-tuning through supervised training, achieving an \fscore \ of 98.04\% $\pm$ 1.09 when identifying cancerous thyroid samples.}

\begin{document}
\thispagestyle{myheadings}
\pagestyle{myheadings}
\markright{\tt Proceedings of CIBB 2019}%check year

\section{\bf Scientific Background}\label{sec:intro}
%\section{\bf Introduction}\label{sec:intro}

\subsection{\bf \it Biological Background}

Cancer can be seen as a collection of diseases, where all are characterized by abnormal and non-stopping cell growth, potentially spreading to surrounding tissues. In 2018, cancerous conditions were the second leading cause of death, worldwide, being responsible for 9.6 million deaths, where approximately 70\% occurred in developing countries~\cite{CancerWHO}. Gene expression is the phenotypic manifestation of a gene or genes by the processes of genetic transcription and translation~\cite{GeneExpr}. Its analysis can help understand the molecular cancer basis better, that can directly influence the prognosis, diagnosis, and treatment of such conditions. The main cancer genomics projects, such as The Cancer Genome Atlas (TCGA) \footnote{\url{https://tcga-data.nci.nih.gov/}} and the International Cancer Genome Consortium\footnote{\url{https://icgc.org}}, try to translate gene expression, by cataloging and profiling through next-generation sequencing thousands of samples across different types of cancers. With more than 50k gene representative features, one can find in these projects genome-wide gene expression assays datasets. It can be a challenge working with this type of data, due to (1) a small number of examples, (2) lack of balance distribution between classes, and (3) potential underlying noise, caused by eventual technical and biological covariates~\cite{Kukurba2015}.

%\section{\bf Related Work}\label{sec:baseline}
\subsection{\bf \it Technical Background}\label{subsec:techB}

Given its high mortality rate, it is crucial to correctly and accurately classify this type of diseases. This need has led many research groups to experiment and study the application of Machine Learning algorithms, as an aim to model the progression and the treatment of cancerous conditions~\cite{KOUROU20158}.

Xie \emph{et al.} developed a predictive model based on a combination of a Multilayer Perceptron and Stacked Denoising Autoencoder (MLP-SAE), to assess how good genetic variants will contribute to gene expression changes~\cite{Xie2017}. The described model is composed of 4 layers (one for the input, two hidden layers from two autoencoders (AEs), and one for the output), with the Mean Squared Error (MSE) as the loss function. Firstly, the authors trained the AEs with a stochastic gradient descent algorithm to later use them on the multilayer perceptron training phase (\emph{i.e.} they use the AEs as weight initialization). The authors used cross-validation to select the optimal model to subsequently (1) compare its performance with the Lasso and Random Forest methods, and (2) evaluate its performance when predicting the gene expression values, on an independent dataset. The authors concluded that the MLP-SAE model: (1) with an MSE of 0.2890 outperformed both previously referred methods (0.2912 and 0.2967, accordingly), and (2) can capture the changes in gene expression quantification.

\cite{Teixeira2017} describes the analysis of the combination of different methods of unsupervised feature learning --- \emph{viz.} Principal Component Analysis (PCA), Kernel Principal Component Analysis (KPCA), Denoising Autoencoder (DAE), and Stacked Denoising Autoencoder --- with different sampling methods for classification purposes. The authors focused on studying the influence of the input nodes on the reconstructed output of the AEs, when feeding these combinations results to a \emph{shallow} artificial network, for distinguishing papillary thyroid carcinoma from healthy samples. In 5-fold cross-validation, the combination of a SMOTE and Tomek links, with a KPCA, was the one with the best overall performance, with a mean \fscore \ of 98.12\%. Notwithstanding, Teixeira \emph{et al.} preferred the usage of a DAE, affirming it yielded similar results (though with a mean \fscore \ of 94.83\%). 

In~\cite{FerreiraBIBM2018}, the authors developed a methodology for the detection of papillary thyroid carcinoma. Ferreira \emph{et al.} studied and compared the performance of a deep neural network classifier architecture, where they used autoencoders (AEs) as a weight initialization method. The AEs were pre-trained to minimize the reconstruction error and  subsequently used to initialize the weights of the top layers of the classification network, with two different strategies: (1) Just the encoding layers, and (2) All the pre-trained AE. 6 types of AEs were used: Basic AE, Denoising AE, Sparse AE, Denoising Sparse AE, Deep AE, and Deep Sparse Denoising AE. Sampling, data augmentation, and normalization techniques when pre-processing the data were not applied. To evaluate and support the results, the authors used stratified 5-fold cross-validation to split the data into training and validation partitions, providing 4 different metrics: Loss, Precision, Recall, and \fscore. Their best result was the combination of unsupervised feature learning through a single-layer Denoising AE, followed by its complete import into the classification network, and subsequent fine-tuning through supervised training, achieving an \fscore \ of 99.61\%, with a variance of 0.54.

\section{\bf Materials and Methods}\label{sec:method}

\subsection{\bf \it The Data}\label{subsec:data}

We used 3 different RNA-Seq datasets from The Cancer Genome Atlas (TCGA), each one representing a type of cancer: thyroid, skin, and stomach. A small sample of one of the datasets is shown in Table \ref{table:1}. All three datasets are composed of the same 20442 features (genes). Each feature represents one certain gene, where the cell values in the table represent the expression of that gene, for a certain sample. The thyroid cancer dataset has 509 examples, the skin cancer dataset 472 and the stomach cancer dataset 415.

Each dataset is processed separately. We start removing, for each one, the features that had the same value for all the instances in the dataset. When a value is constant for all the examples, there is no entropic value (\emph{i.e.}, it is not possible to infer any information). We then imputed the missing values (\emph{NA}'s, as shown in Table~\ref{table:1}) with the average value of its respective column, and added (to each one) a column \emph{Label} to match each instance to its type of cancer. Our goal is to distinguish different types of cancer, so we assign a positive value (1) to the class we want to predict, and 0 to the remaining ones: when training the model to detect thyroid cancer, all thyroid examples are labeled as 1 and the skin and stomach instances as 0. Respectively, when training to detect skin/stomach cancer, all skin/stomach examples are labeled as 1 and the remaining two types of cancer's instances are labeled as 0. However, after this process, it is not guaranteed (and actually quite unlikely) that the same features will be removed in the 3 cancer datasets. Thus, when merging the 3 sets of data, we only use their intersection, so that the different types of cancer are represented by the same features. After the full data pre-processing, the final dataset has 18321 feature columns and 1396 examples (36\% of thyroid cancer, 34\% of skin cancer, and 30\% of stomach cancer).

\begin{table*}[t]
\vspace{3mm}
\centering
\caption{An example of 5 samples of the thyroid dataset. The header line represents the names of the genes and column values represent its expression for each sample. \emph{NA} means that a value is missing, for that gene, and sample.}
\resizebox{\textwidth}{!}{
\begin{tabular}{llrrrrrrr}
\toprule
 & sampleId & UBE2Q2P2\_100134869 & HMGB1P1\_10357 & LOC155060\_155060&  ... & ZZZ3\_26009 &  TPTEP1\_387590 & AKR1C6P\_389932 \\
\midrule
0 &  TCGA-4C-A93U-01 &              -1.6687 &             NA &                NA &    ... &   -0.9478 &          -1.3739 &               NA \\
1 &  TCGA-BJ-A0YZ-01 &              -1.1437 &             NA &                NA &    ... &   -0.4673 &          -0.0166 &               NA \\
2 &  TCGA-BJ-A0Z0-01 &              -0.9194 &             NA &                NA &    ... &    2.1918 &          -1.5856 &               NA \\
3 &  TCGA-BJ-A0Z2-01 &               1.1382 &             NA &                NA &    ... &    1.5512 &          -1.5897 &               NA \\
4 &  TCGA-BJ-A0Z3-01 &              -0.3333 &             NA &                NA &    ... &    0.4926 &          -1.3379 &               NA \\
\bottomrule
\end{tabular}}
\label{table:1}
\end{table*}

\subsection{\bf \it Autoencoders}

An AE is a neural network that aims to reproduce its input~\cite{Rumelhart:1986}. Let $f$ and $g$ correspond to the encoding and decoding functions of the AE, parameterized on $\theta_e$ and $\theta_d$ respectively, where $\theta = \theta_e \cup \theta_d$, $L$ being an appropriate loss function, and $J$ the cost function to be minimized. In its learning process, an AE tries to find the value for $\theta$ that leads to the minimal value of function $J(\theta,X) = L(X, g_{\theta_d}(f_{\theta_e}(X))$, assigning a penalty to the reconstruction of the input $\hat{X}=g_{\theta_d}(f_{\theta_e}(X))$ when it is distinct from the original data $X$~\cite{Goodfellow2016DL}. 

In this work we chose to only compare the performance of the AE that had the best result in~\cite{FerreiraBIBM2018} (DAE) as weight initialization of a classification architecture, studying two different approaches for weight initialization and two different strategies for embedding the AE layers. A DAE~\cite{Vincent:2008} is a type of AE that tries to preserve the input's information, undoing the effect of a corruption process applied to the input of the AE, by $J(\theta,X) = L(X, g_{\theta_d}(f_{\theta_e}(\tilde{X})))$, where $\tilde{X}$ is a copy of the input $X$, corrupted by some form of \emph{noise}~\cite{Goodfellow2016DL}. In our case, we apply a \emph{Dropout} layer, directly after the input layer as a form of Bernoulli Noise, where 10\% of the connections are randomly deleted. The hidden encoding layer size is 128. 

\subsection{\bf \it Methodology}\label{sec:pipeline}

As in \cite{FerreiraBIBM2018}, our experiment consists in the performance assessment of a deep neural network classifier architecture, where we vary its top layers. However, we aim to identify 3 distinct types of cancer, instead of distinguishing cancerous from healthy samples. We pre-train the autoencoders to minimize the reconstruction error and subsequently use them to initialize the top layers weights of the classification network, with two different strategies: \textbf{(1)} Just the encoding layers, and \textbf{(2)} All the pre-trained autoencoder.

Each architecture is thus trained to classify the input data as either \emph{thyroid}, \emph{skin} or \emph{stomach}, accordingly to the type of cancer. We use the same architecture as Ferreira \emph{et. al} and, given that such architecture was build for a binary classification task, we decided to adapt this multi-label classification problem to a ``binary label'' one: for a type of cancer $C$, we train the model to detect $C$ and not $C$, instead of detecting cancer and healthy samples. Besides the top layers imported from the AE, the classification region of the full network starts with a Batch Normalization layer, and proceeds with two Fully Connected layers using Rectified Linear Unit (ReLU) activation; the last one --- prediction layer --- is a single neuron layer with a Sigmoid non-linearity.

\begin{figure*}[h]
\centering
  \begin{subfigure}[t]{0.32\textwidth}
    \includegraphics[width=\textwidth]{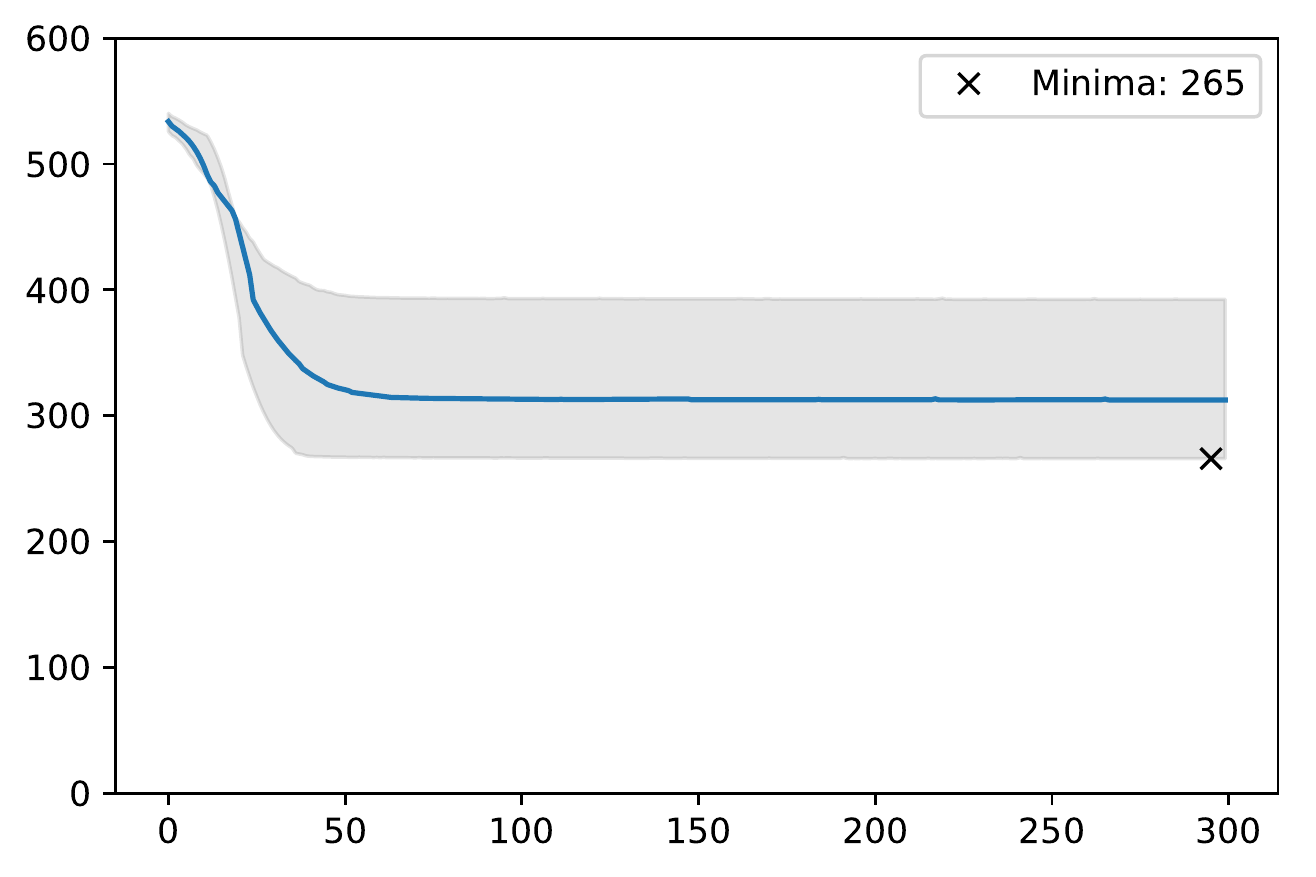}
    \caption{Basic AE}
    \label{fig:3a}
  \end{subfigure}
  \begin{subfigure}[t]{0.32\textwidth}
    \includegraphics[width=\textwidth]{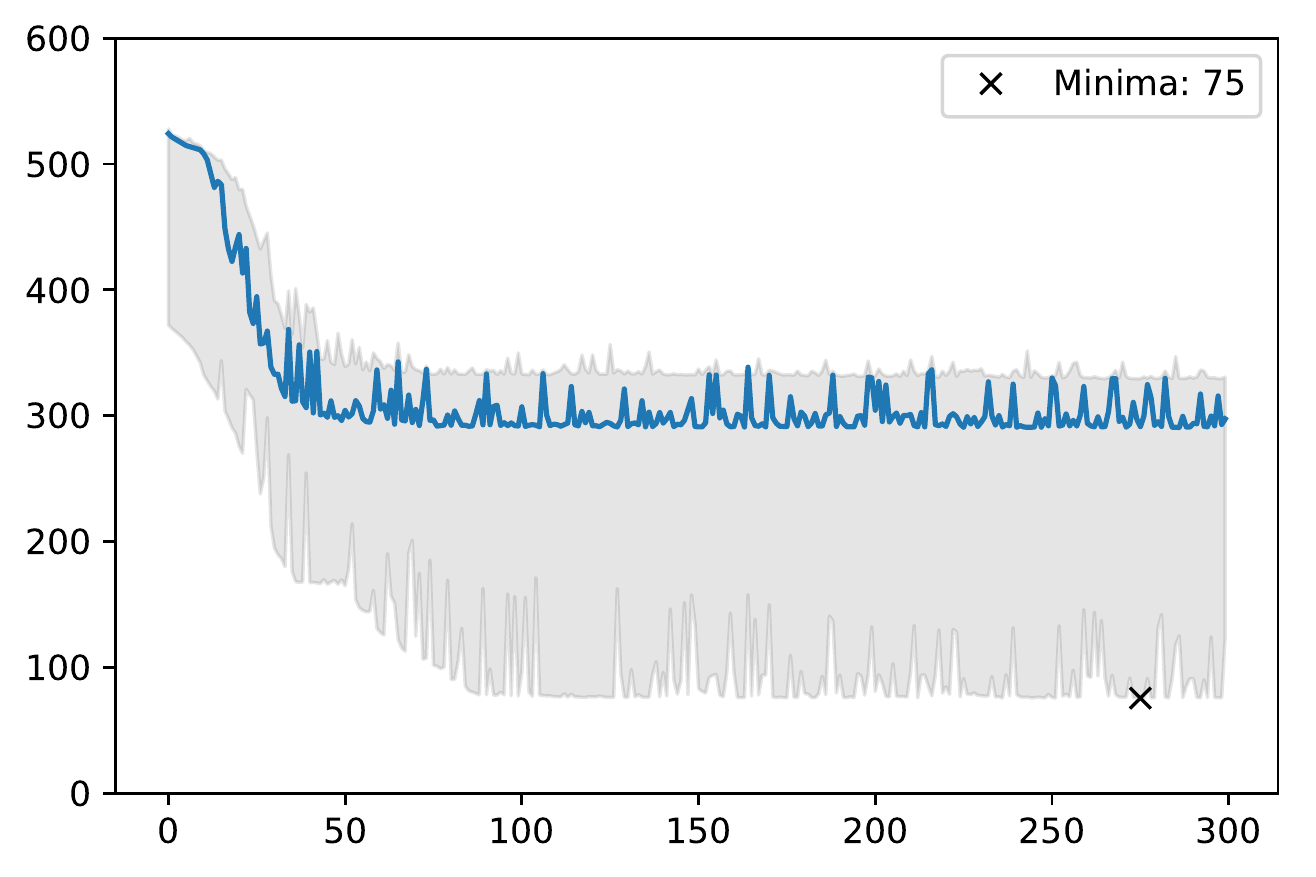}
    \caption{Denoising AE}
    \label{fig:3b}
  \end{subfigure}
    \begin{subfigure}[t]{0.32\textwidth}
    \includegraphics[width=\textwidth]{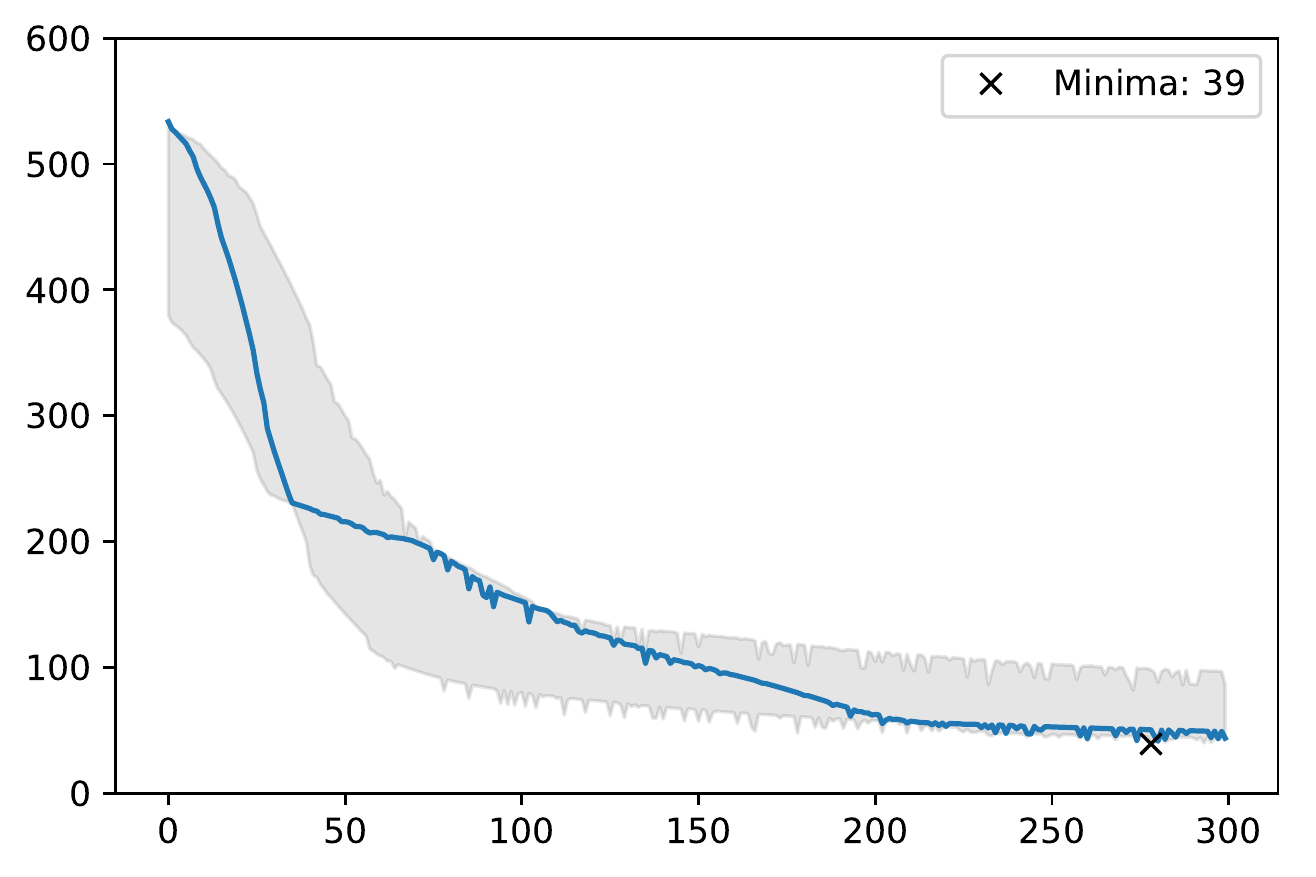}
    \caption{Sparse AE}
    \label{fig:3c}
 	\end{subfigure}
  	\begin{subfigure}[b]{0.32\textwidth}
    \includegraphics[width=\textwidth]{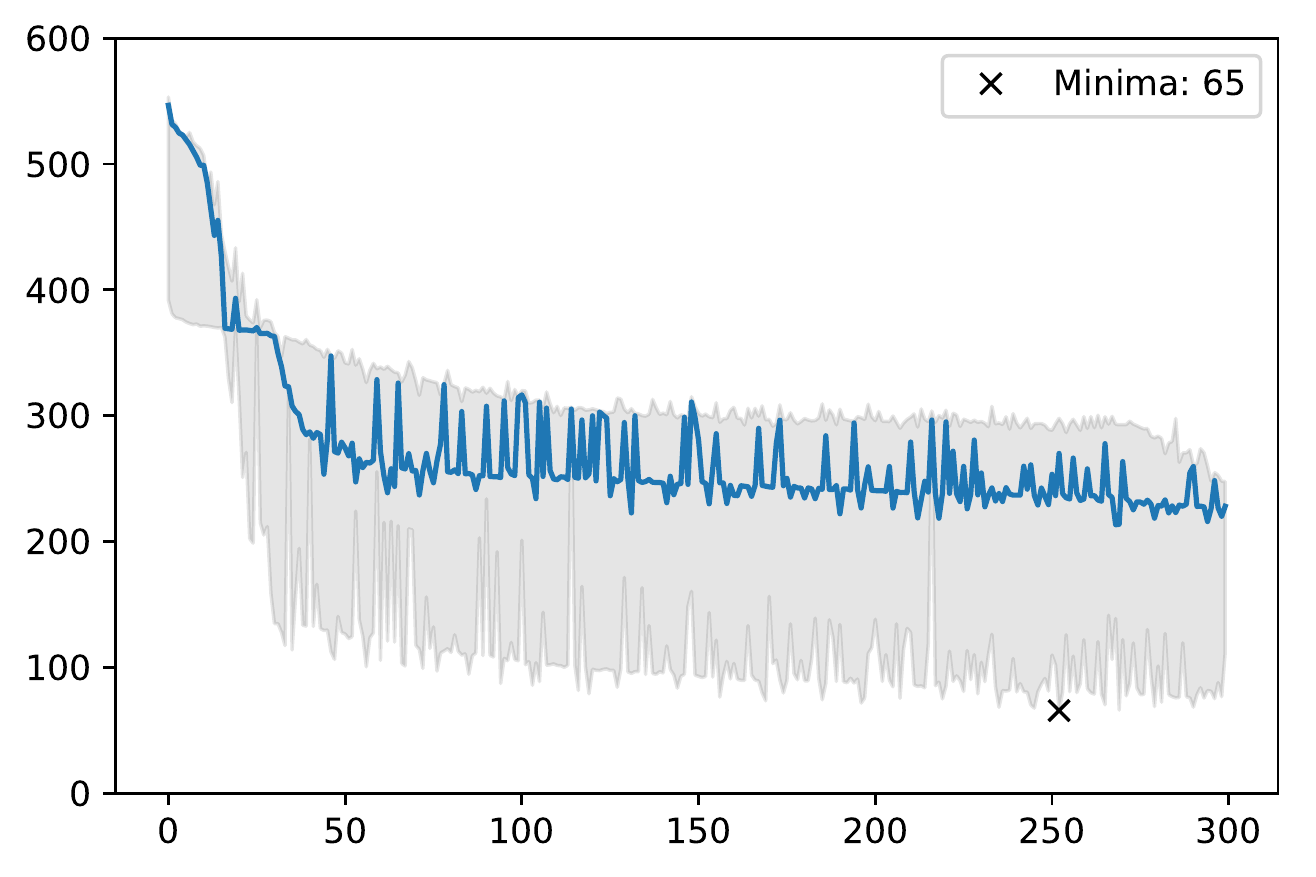}
    \caption{Denoising Sparse AE}
    \label{fig:3d}
  \end{subfigure}
  \begin{subfigure}[b]{0.32\textwidth}
    \includegraphics[width=\textwidth]{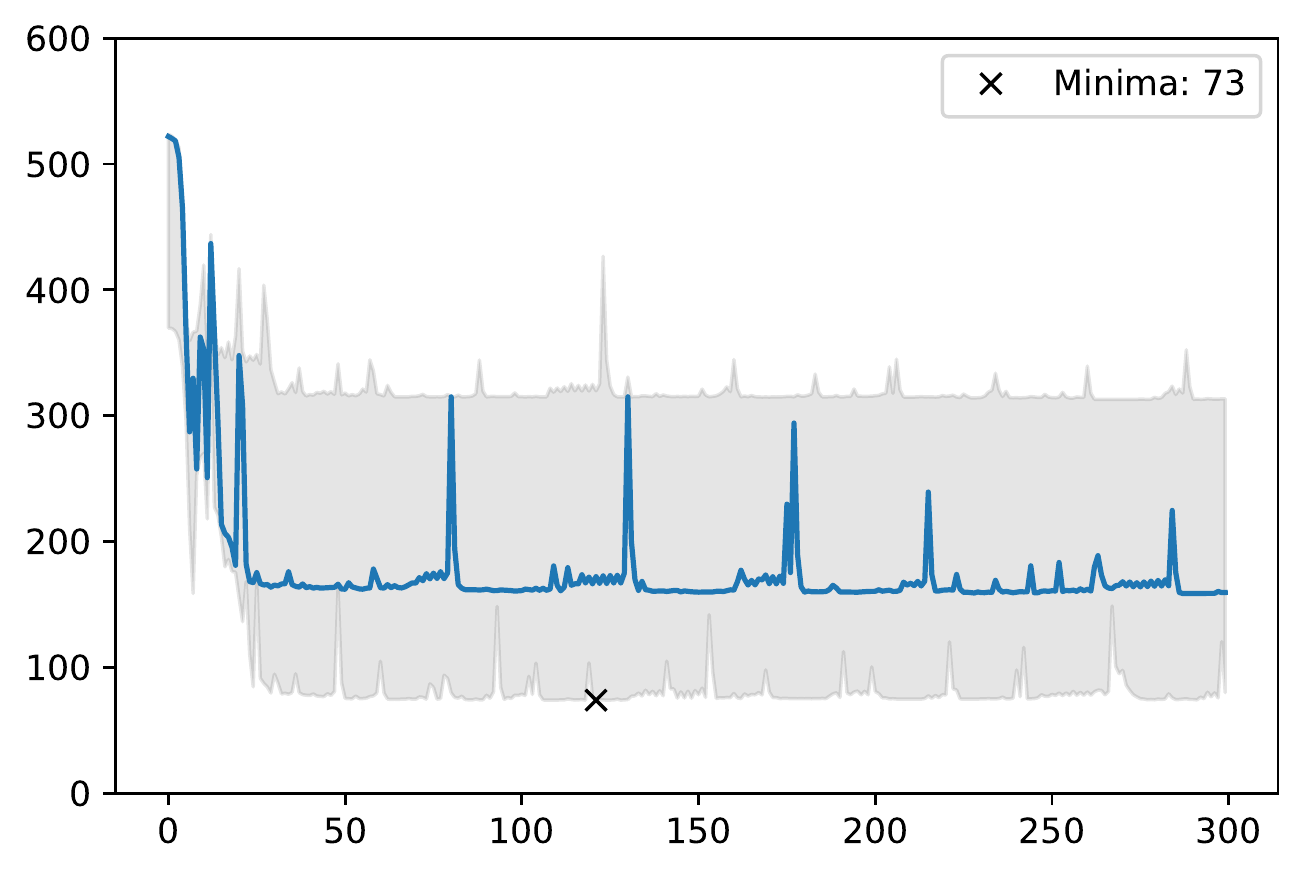}
    \caption{Deep AE}
    \label{fig:3e}
  \end{subfigure}
    \begin{subfigure}[b]{0.32\textwidth}
    \includegraphics[width=\textwidth]{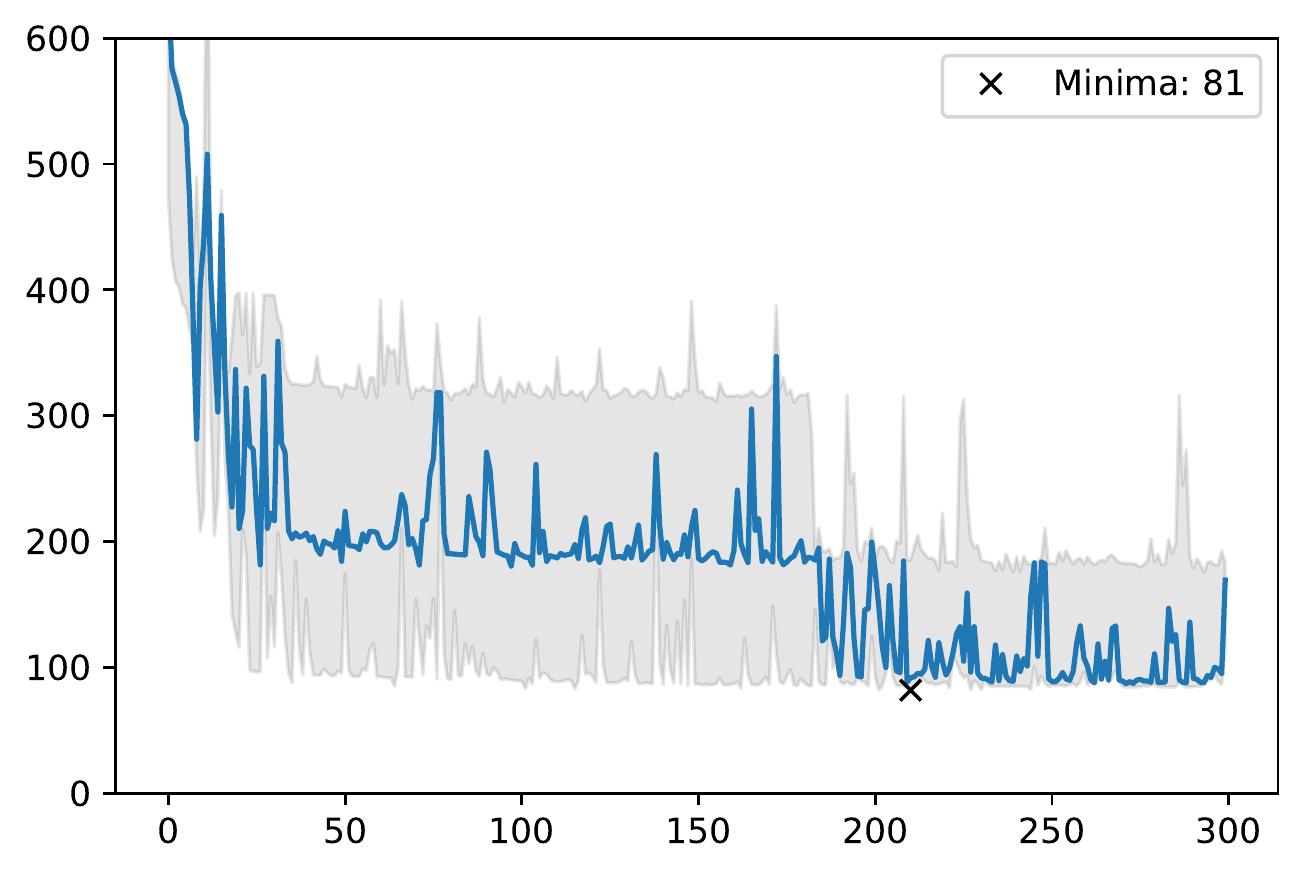}
    \caption{Deep Sparse Denoising AE}
    \label{fig:3f}
  \end{subfigure}
  \caption{Loss values on the training set for the 300 epochs of autoencoder training, with corresponding minimas. The \emph{x} axis represents the number of epochs, and the \emph{y} axis the loss value. The grey area represents the variance of the loss value.}
  \label{fig:3}
\end{figure*}

\begin{figure*}[h]
\centering
  \begin{subfigure}[t]{0.32\textwidth}
    \includegraphics[width=\textwidth]{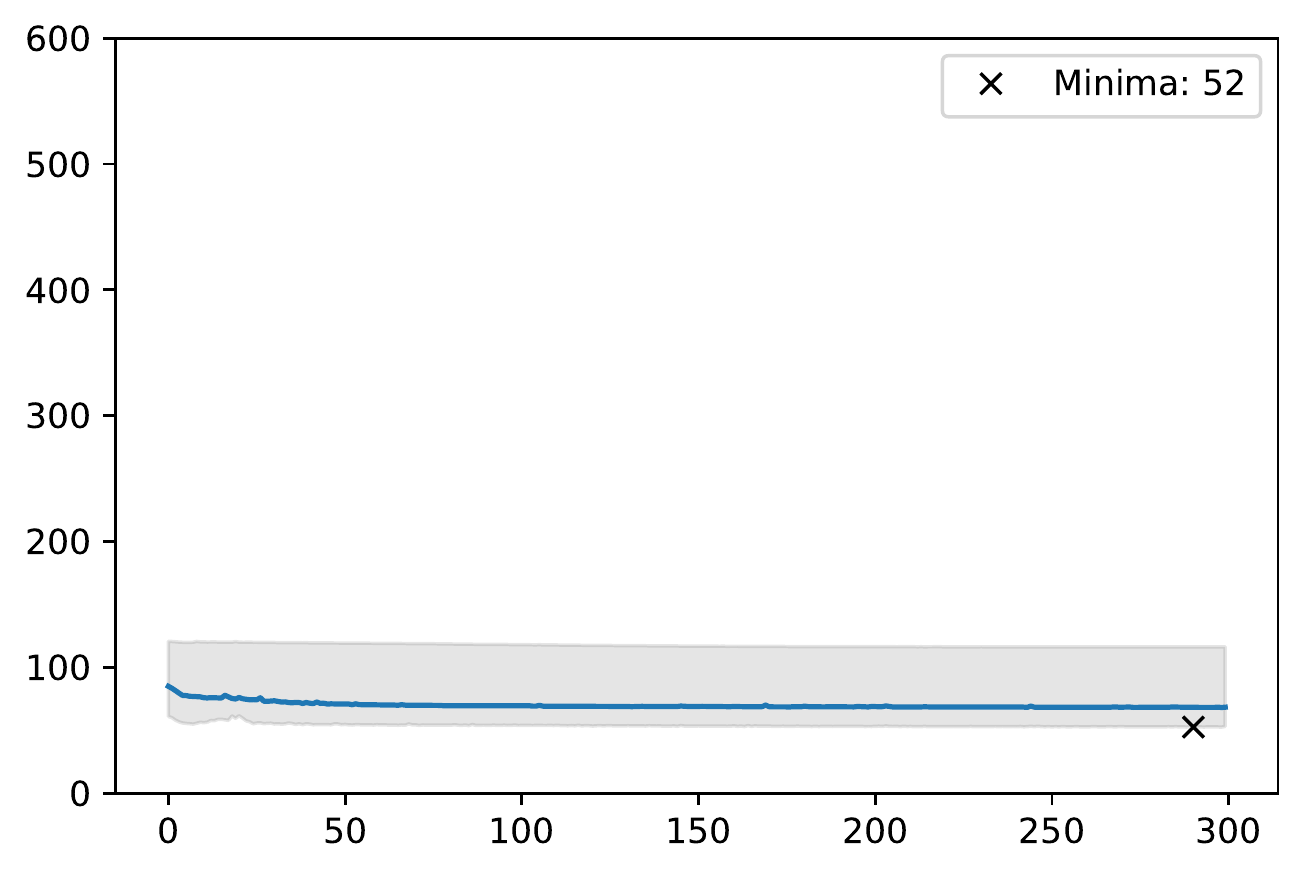}
    \caption{Basic AE}
    \label{fig:2a}
  \end{subfigure}
  \begin{subfigure}[t]{0.32\textwidth}
    \includegraphics[width=\textwidth]{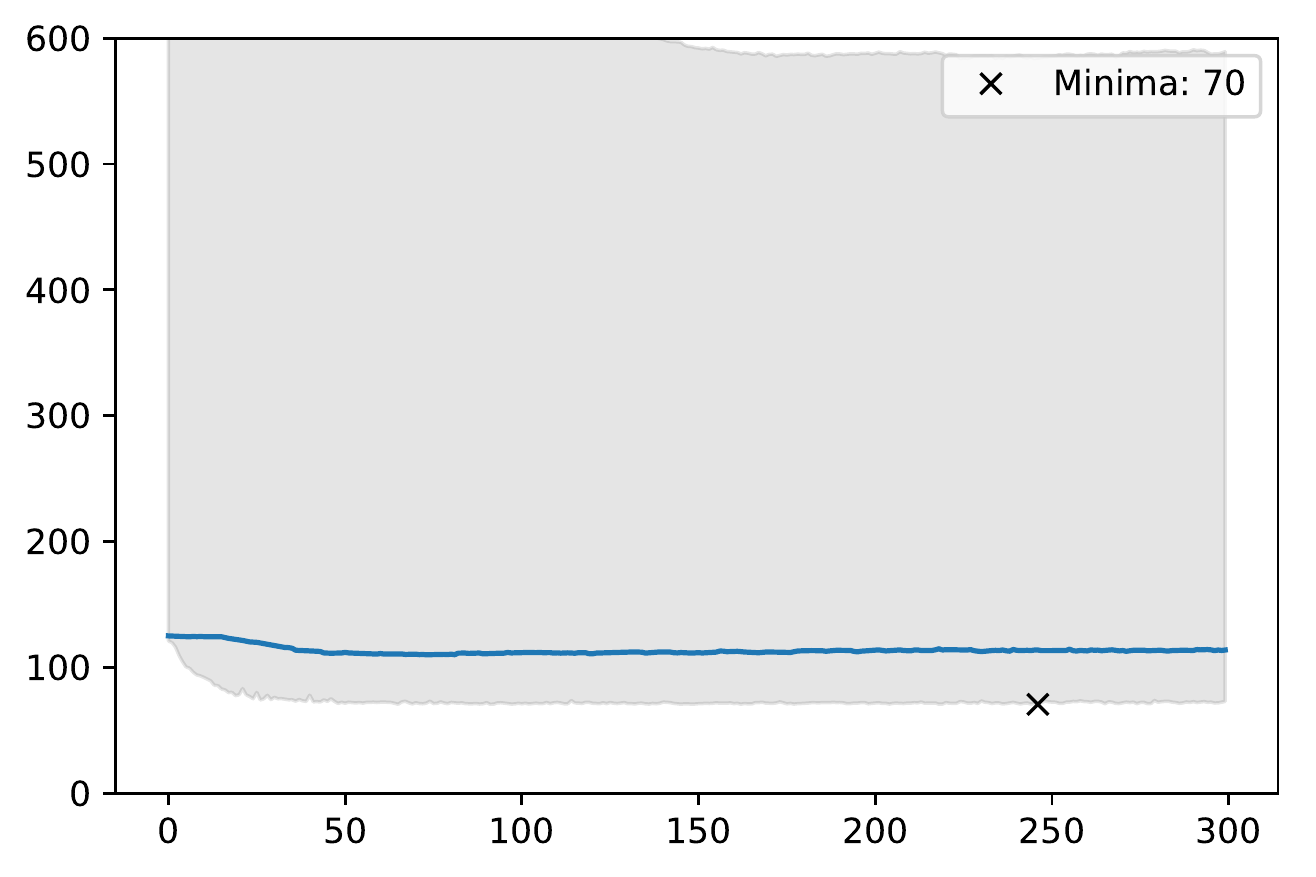}
    \caption{Denoising AE}
    \label{fig:2b}
  \end{subfigure}
    \begin{subfigure}[t]{0.32\textwidth}
    \includegraphics[width=\textwidth]{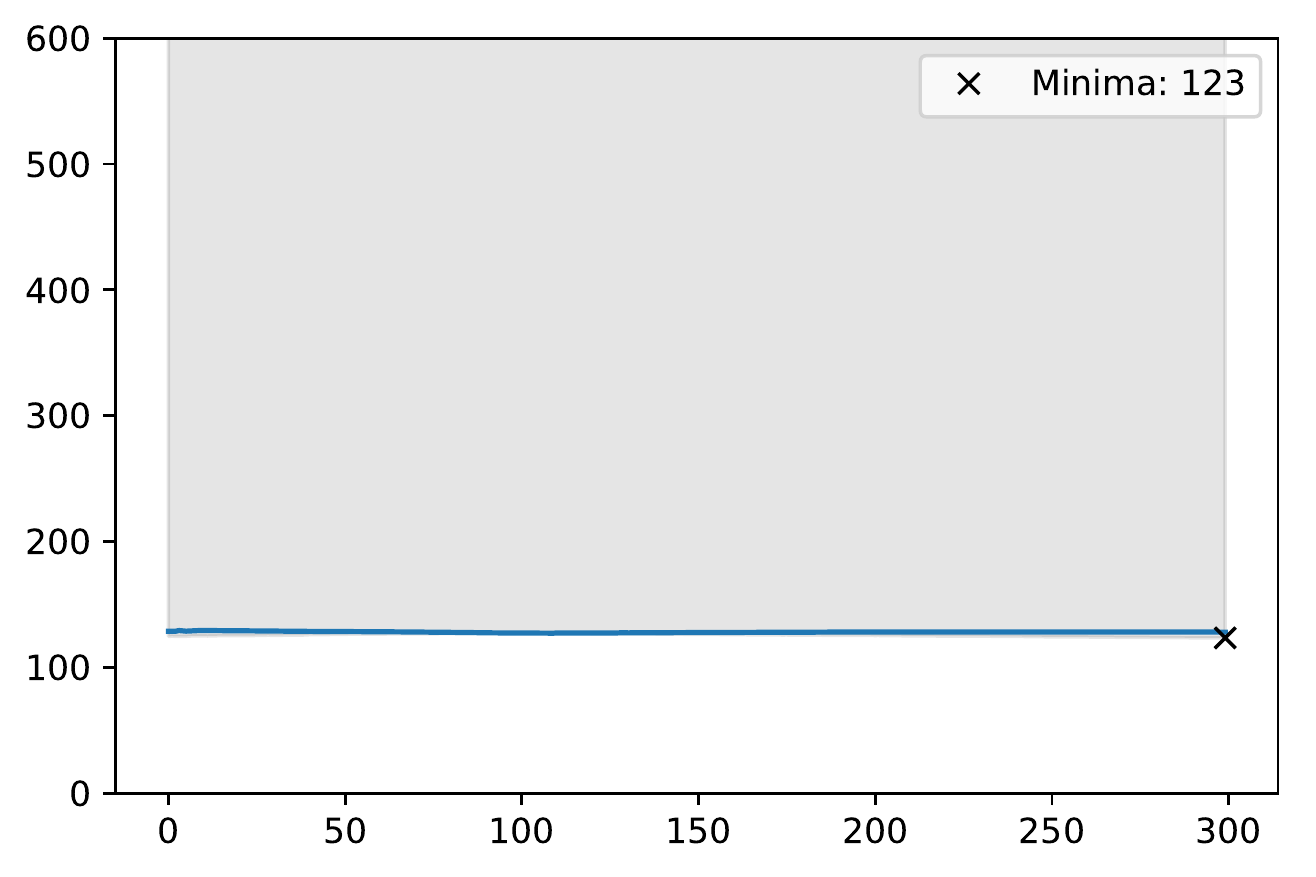}
    \caption{Sparse AE}
    \label{fig:2c}
 	\end{subfigure}
  	\begin{subfigure}[b]{0.32\textwidth}
    \includegraphics[width=\textwidth]{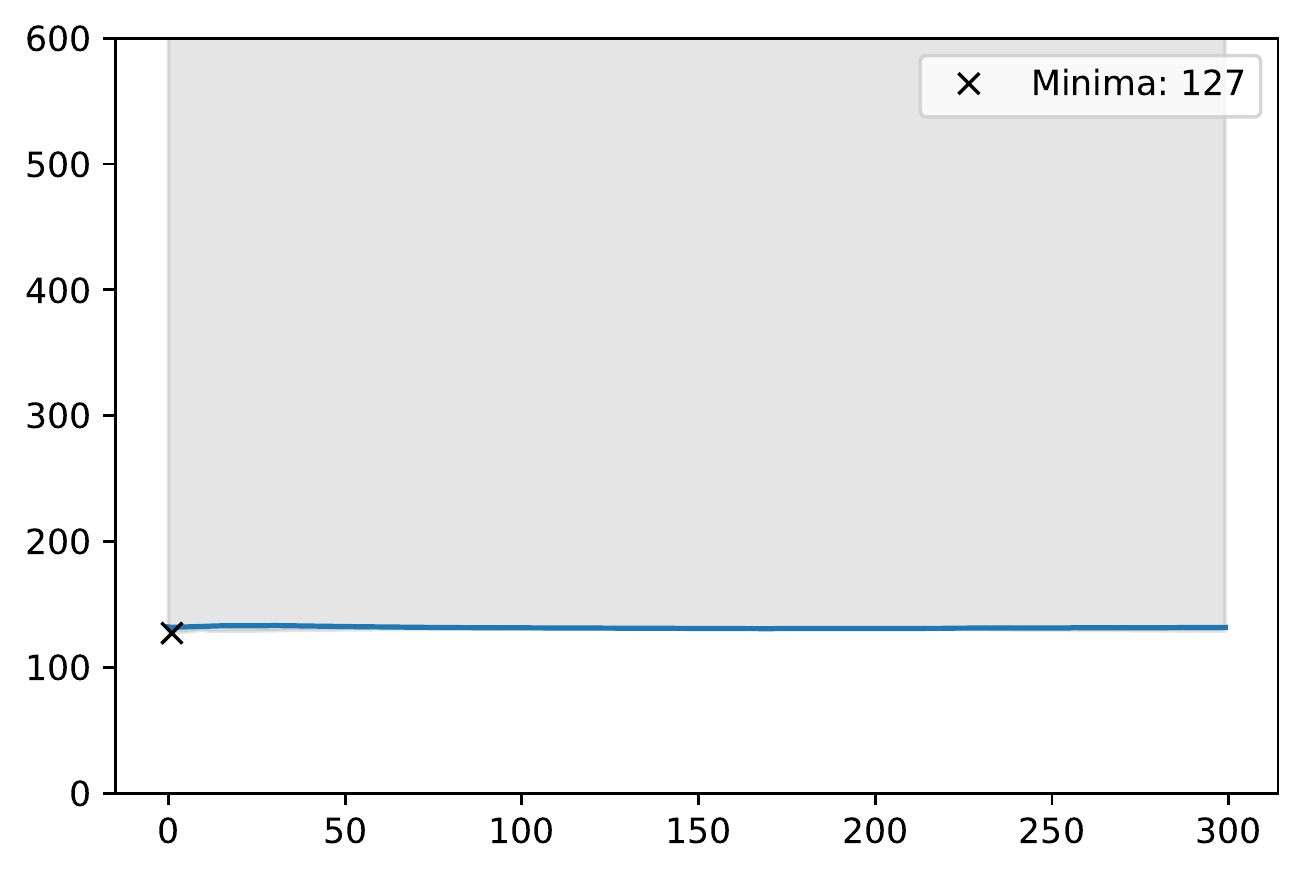}
    \caption{Denoising Sparse AE}
    \label{fig:2d}
  \end{subfigure}
  \begin{subfigure}[b]{0.32\textwidth}
    \includegraphics[width=\textwidth]{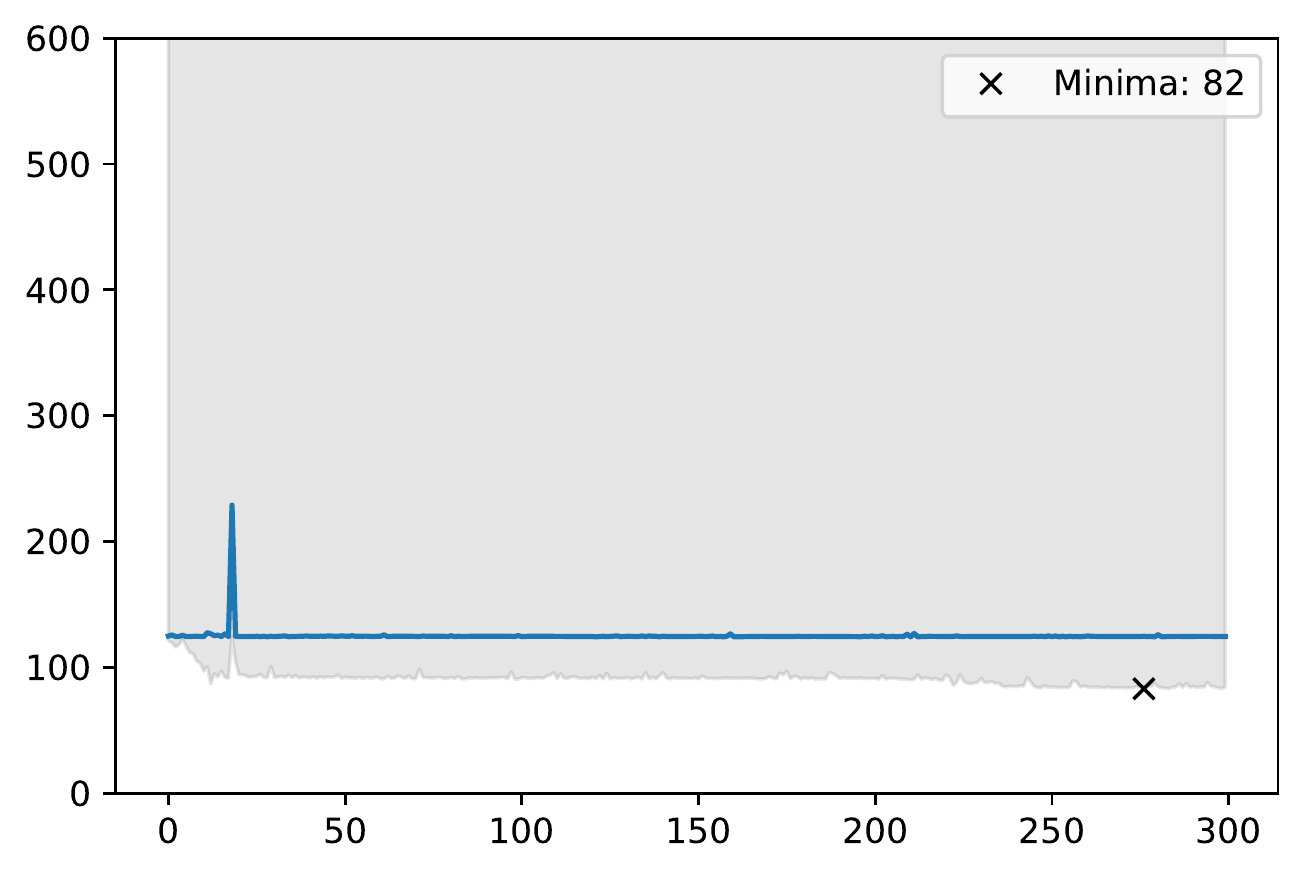}
    \caption{Deep AE}
    \label{fig:2e}
  \end{subfigure}
    \begin{subfigure}[b]{0.32\textwidth}
    \includegraphics[width=\textwidth]{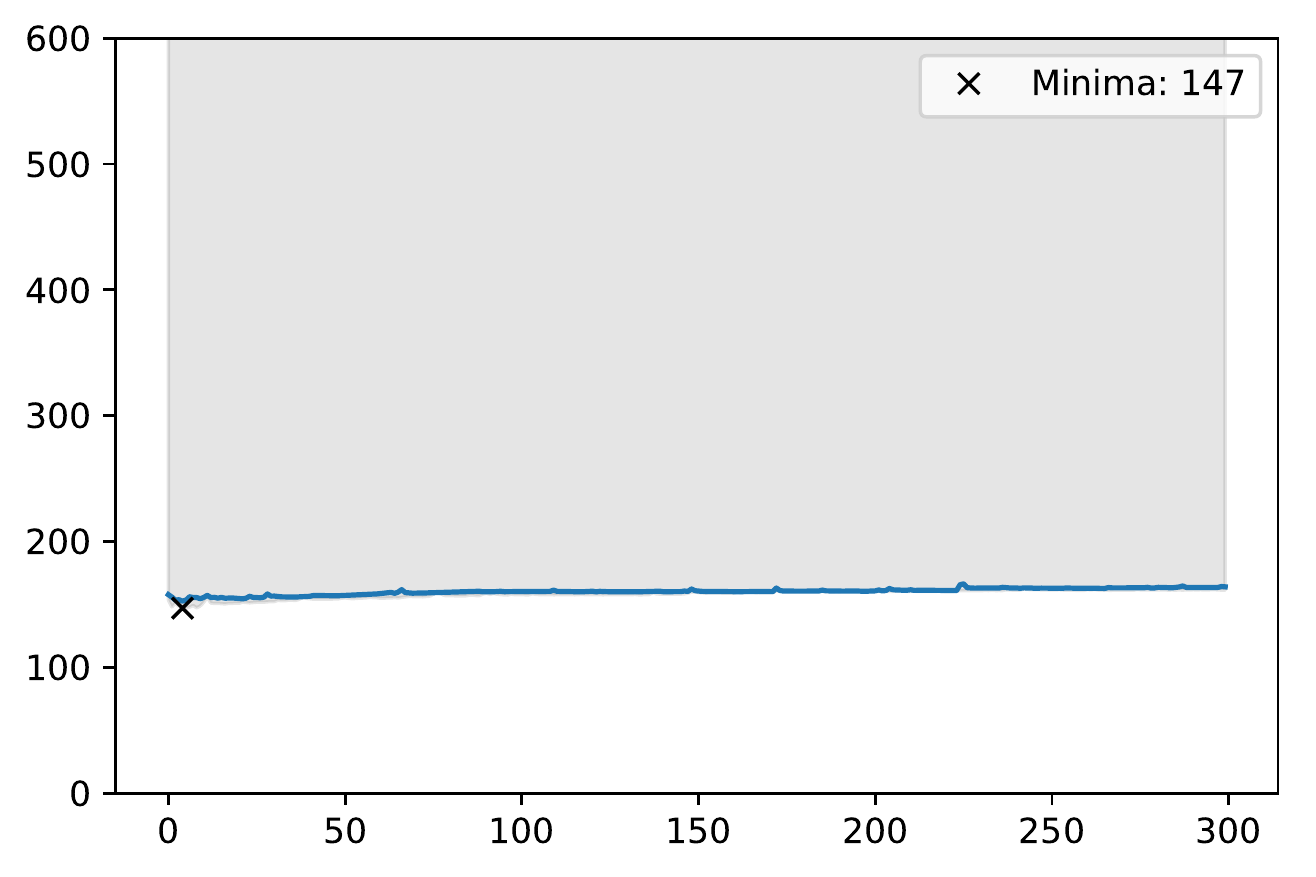}
    \caption{Deep Sparse Denoising AE}
    \label{fig:2f}
  \end{subfigure}
  \caption{Loss values on the validation set for the 300 epochs of autoencoder training, with corresponding minimas. The \emph{x} axis represents the number of epochs, and the \emph{y} axis the loss value. The grey area represents the variance of the loss value.}
  \label{fig:2}
\end{figure*}

\subsection{\bf \it Evaluation}

In order to ensure and provide statistical evidence, we use stratified 5-fold cross-validation. The DAE and classifier are trained during 100 and 300 epochs, respectively, with a batch size of 500. The loss of the classifier model is calculated by the \emph{binary cross-entropy}~\cite{Goodfellow2016DL}, and trained using an \emph{adam} optimizer. We then evaluate its performance through 4 additional metrics: Accuracy, Precision, Recall, and \fscore also for the training and the validation sets.

\section{\bf Results and Discussion}\label{sec:disc}

\textbf{One tends to assume that the previously described methodology can be generalized to other datasets and problems:} Importing the complete pre-trained DAE to the upper layers of the classification architecture and allowing subsequent \emph{fine-tuning} achieved the best overall performance, with an \fscore \ of 98.04\% (when detecting thyroid cancer), a result that is quite close to the overall best of 99.61\% reported in \cite{FerreiraBIBM2018}. However, for both detection of skin and stomach cancers, the best-achieved result was, respectively, 97.81\% ($\pm$ 1.76) and 97.54\% ($\pm$ 1.25), where the combination differs only on the DAE layers that are embedded into the classifier (only the encoding layers). We may assume that this methodology can generalize to other types of data.

\begin{table*}[t]
\vspace{3mm}
\caption{Performance comparison of the classifier. We are only importing the top layers from a DAE since it was the AE that led to better results in~\cite{FerreiraBIBM2018} (where the best overall result was the combination of a Complete DAE, with Approach B, achieving an \fscore \ of 99.61\% $\pm$ 0.54). \emph{T} represents thyroid cancer detection, \emph{Sk} skin cancer detection, and \emph{St} stomach cancer detection. When measuring loss, lower is better. For all the other metrics, higher is better. All the values presented are the average value of a 5-fold cross-validation, at the validation set, by selecting the best performing model according to its \emph{\fscore}.}
\centering
    \resizebox{\textwidth}{!}{\begin{tabular}{llccccc|ccccc}
    \toprule
    \multirow{2}{*}{} &
        \multicolumn{5}{c|}{Fixed Weights (Approach A)} &
        \multicolumn{5}{c}{Fine-Tuning (Approach B)} \\
    \cmidrule{2-6} \cmidrule{7-11}
    Top Layers (DAE) & Loss & Accuracy (\%) & Precision (\%) & Recall (\%) & \fscore \ (\%) & Loss & Accuracy (\%) & Precision (\%) & Recall (\%) & \fscore \ (\%)\\
    \midrule
    T: Encoding Layers  & 0.309 $\pm$ 0.37 & 89.12\% $\pm$ 2.44 & 82.80\% $\pm$ 4.36 & 88.81\% $\pm$ 3.69 & 85.63\% $\pm$ 3.08 & 0.117 $\pm$ 0.50 & 98.35\% $\pm$ 0.86 & 96.06\% $\pm$ 2.13 & 99.61\% $\pm$ 0.54 & 97.79\% $\pm$ 1.13 \\
    T: Complete AE  & 0.375 $\pm$ 0.16 & 92.41\% $\pm$ 2.69 & 86.82\% $\pm$ 6.09 & 93.91\% $\pm$ 1.28 & \textbf{90.11\% $\pm$ 3.09} & 0.662  $\pm$ 0.49 & 98.57\% $\pm$ 0.80 &  97.88\% $\pm$ 1.83 & 98.23\% $\pm$ 1.76 &\textbf{ 98.04\% $\pm$ 1.09} \\ 
    \midrule
    Sk: Encoding Layers  & 0.346 $\pm$ 0.46 & 88.82\% $\pm$ 2.36 & 91.19\% $\pm$ 3.75 & 74.34\% $\pm$ 8.33 & 81.60\% $\pm$ 4.91 & 0.545 $\pm$ 0.02 & 98.57\% $\pm$ 1.13 & \textbf{100.00\% $\pm$ 0.00} & 95.76\% $\pm$ 3.37 & 97.81\% $\pm$ 1.76 \\
    Sk: Complete AE  & 0.482 $\pm$ 0.07 & 91.33\% $\pm$ 2.55 & 88.02\% $\pm$ 6.69  & 86.65\% $\pm$ 4.03 & 87.16\% $\pm$ 3.53 & 0.893 $\pm$ 0.03 & 98.14\% $\pm$ 0.46 & 98.27\% $\pm$ 0.59 & 96.19\% $\pm$ 0.94 & 97.22\% $\pm$ 0.70 \\ 
    \midrule
    St: Encoding Layers  & 0.431 $\pm$ 0.06 & 83.16\% $\pm$ 4.14 & 90.70\% $\pm$ 5.96 & 47.95\% $\pm$ 13.32 & 62.01\% $\pm$ 12.99 & 0.590 $\pm$ 0.03 & 98.57\% $\pm$ 0.72 & 99.25\% $\pm$ 0.68 & 95.90\% $\pm$ 2.19 & 97.54\% $\pm$ 1.25 \\
    St: Complete AE  & 0.465 $\pm$ 0.12 &  89.83\% $\pm$ 2.18  & 85.13\% $\pm$ 3.14 & 80.00\% $\pm$ 9.36 & 82.16\% $\pm$ 4.90 & 0.147 $\pm$ 0.10 & 97.49\% $\pm$ 1.81 & 97.95\% $\pm$ 1.50 & 93.49\% $\pm$ 5.15 & 95.63\% $\pm$ 3.23 \\
    \bottomrule
    \end{tabular}}
    \label{table:2}
\end{table*}

\textbf{Fine-tuning (Approach B) leads to better results than fixing the weights (Approach A):} In~\cite{FerreiraBIBM2018}, the authors claimed that their results cannot support that Approach B gave better results than Approach A. However, with our data, it is clear that fine-tuning the weights of the top layers leads to better results, by a margin of 10 -- 20\%, when considering the \fscore~metric, as one can see in Table~\ref{table:2}.

\textbf{There is not enough evidence to support the assumption that the overall usage of AEs seem to capture the most relevant information for the task:} Although our overall best was close to the overall best of the previously referred work, there is a big difference between the two approaches of weight initialization when experimenting our data. Also, there is a big divergence when analyzing the AEs curves in the train and validation phases, as it is observable in Figure~\ref{fig:2} and Figure~\ref{fig:3}. One may assume that the AEs learning process is being compromised given that (1) in some cases, in the validation phase (for example the DSAE -- Figure~\ref{fig:2d} -- and the DSDAE -- Figure~\ref{fig:2f}), the minima is found too early and (2) the data split in the cross-validation may have influence on the learning process.

\section{\bf Conclusions}~\label{sec:concl}

In this work, we havecompared the performance of a Denoising Autoencoder (DAE) as an unsupervised initialization method for deep classification neural networks applied to a cancer \emph{vs.} cancer classification task. For that, we have used the methodology described in~\cite{FerreiraBIBM2018}: we combined a DAE with two different approaches when training the classification architecture: (a) by fixing the imported weights, and (b) by allowing them to be fine-tuned during supervised training. We studied two different strategies for embedding the DAE into the classification network: (1) using the encoding layers as weight initialization, and (2) using the complete AE, \emph{i.e.}, both the encoding and decoding layers.

Taking Ferreira \emph{et al.} as a reference model, we think that it may be possible to generalize the methodology to other datasets and problems. Importing a complete pre-trained DAE to the top layers of the classifier (Strategy 2), followed by fine-tuning (Approach B), when detecting thyroid cancer, achieved the best overall results, with an \fscore \ of 98.04\% $\pm$ 1.09. Fine-tune led to better results, boosting the results between 10 and 20\% in the \fscore~metric. Contrary to the results obtained in the mentioned previous work, there is not enough evidence to support the assumption that the overall usage of AEs seems to capture the most relevant information for the task, in this problem.

%\section*{\bf Acknowledgments}
%Example of the Acknowledgments section.

\bibliographystyle{apalike}
{\fontsize{10}{10}\selectfont
\bibliography{bibl}}

\end{document}